\documentclass[runningheads]{llncs}
\usepackage[T1]{fontenc}
\usepackage{graphicx}
\usepackage{xcolor}
\usepackage{caption}
\usepackage{subcaption}
\usepackage{multirow}
\usepackage{hyperref}
\usepackage{fancyhdr}

\usepackage{svg}
\usepackage{orcidlink}

\newcommand\tabref[1]{Table \ref{#1}}
\newcommand\figref[1]{Figure \ref{#1}}
\newcommand\subfigref[1]{Figure (\subref{#1})}
\newcommand\secref[1]{Section \ref{#1}}

\newcommand\methodours{Ours}
\newcommand\methodvlad{NetVLAD}
\newcommand\methodRibeiro{Ribeiro18}
\newcommand\methodrandom{Random}

\newcommand{\STAB}[1]{\begin{tabular}{@{}c@{}}#1\end{tabular}}
\begin{document}

\fancyhead[C]{\footnotesize This paper has been accepted at the 14th International Conference on Computer Vision Systems (ICVS 2023) \url{https://doi.org/10.1007/978-3-031-44137-0_28}}
\fancyfoot[C]{\footnotesize Please cite this paper as:\\ I. Donadi, E. Olivastri, D. Fusaro, W. Li, D. Evangelista, and A. Pretto, Improving Generalization of Synthetically Trained Sonar Image Descriptors for Underwater Place Recognition, 14th International Conference on Computer Vision Systems (ICVS), 2023}

\let\oldmaketitle\maketitle
\renewcommand{\headrulewidth}{0pt} 
\renewcommand{\maketitle}{%
  \oldmaketitle
  \thispagestyle{fancy}
  \fancyhead[RE,RO]{}
\fancyhead[LE,LO]{}
}

\title{Improving Generalization of Synthetically Trained Sonar Image Descriptors for\\Underwater Place Recognition}
\titlerunning{Sonar Image Descriptors for Underwater Place Recognition}

\author{
Ivano Donadi
\orcidlink{0009-0002-2362-066X}
\and
Emilio Olivastri
\orcidlink{0000-0002-0003-5559} 
\and
Daniel Fusaro
\orcidlink{0009-0000-4053-5577}
\and
Wanmeng Li
\orcidlink{0009-0003-1171-5685}
\and
Daniele Evangelista
\orcidlink{0000-0001-8972-9145}
\and
Alberto Pretto
\orcidlink{0000-0003-1920-2887}
}

\authorrunning{Donadi et al.}

\institute{
Department of Information Engineering (DEI), University of Padova, Italy\\
Emails: \email{[donadiivan, olivastrie, fusarodani, liwanmeng, evangelista, alberto.pretto]@dei.unipd.it}
}
\maketitle              
\begin{abstract}

Autonomous navigation in underwater environments presents challenges due to factors such as light absorption and water turbidity, limiting the effectiveness of optical sensors. Sonar systems are commonly used for perception in underwater operations as they are unaffected by these limitations. Traditional computer vision algorithms are less effective when applied to sonar-generated acoustic images, while convolutional neural networks (CNNs) typically require large amounts of labeled training data that are often unavailable or difficult to acquire. 
To this end, we propose a novel compact deep sonar descriptor pipeline that can generalize to real scenarios while being trained exclusively on synthetic data. Our architecture is based on a ResNet18 back-end and a properly parameterized random Gaussian projection layer, whereas input sonar data is enhanced with standard ad-hoc normalization/prefiltering techniques. A customized synthetic data generation procedure is also presented. The proposed method has been evaluated extensively using both synthetic and publicly available real data, demonstrating its effectiveness compared to state-of-the-art methods.


\end{abstract}
\section{Introduction}

Autonomous underwater vehicles (AUVs) represent a key enabling technology for carrying out complex and/or heavy but necessary operations in underwater environments in a totally safe way, such as exploration, assembly, and maintenance of underwater plants. AUVs could perform these tasks in a fully autonomous way, without the need for remote piloting and possibly without the need for a support vessel, with undoubted advantages from an economic, environmental, and personnel safety point of view. 

However, autonomous navigation in underwater environments poses significant challenges due to factors such as light absorption and water turbidity. These factors severely limit the effectiveness of optical sensors like RGB cameras. Due to their immunity to the aforementioned limitations, the primary sensors used for perception in underwater operations are sonar systems.
Sonars operate by emitting acoustic waves that propagate through the water until they encounter an obstacle or are absorbed. Nevertheless, sonar-generated acoustic images (often called \emph{sonar images}) are affected by various sources of noise, including multi-path interference, cross-sensitivity, low signal-to-noise ratio, acoustic shadows, and poor pixel activation. As a result, traditional computer vision algorithms such as handcrafted feature detection and descriptor schemes, which typically perform relatively well
on optical images, are less effective when applied to acoustic images.
On the other hand, convolutional neural networks are capable of learning highly effective features from acoustic images. 
However, to effectively generalize, they typically require a large amount of labeled training data, which is often difficult to obtain due to the lack of large publicly available datasets.
To address this limitation, simulation has emerged as an invaluable tool. 
Through synthetically generated data, it is possible to overcome the lack of available datasets and eliminate the need for manual data annotation.\\

In this work, we address the underwater place recognition problem by using a Forward-Looking Sonar (FLS) sensor, with the goal of improving the autonomous capabilities of underwater vehicles in terms of navigation. Like most place recognition pipelines, our goal is to extract a compact sonar image representation, i.e. an n-dimensional vector.
Such compact representation can then be used to quickly match a query image (representing the current place the autonomous vehicle is in) with a database of descriptors (representing places already seen in the past). This process involves techniques such as nearest neighbor search to extract the most probable matching image. We introduce a new deep-features-based global descriptor of acoustic images that is able to generalize in different types of underwater scenarios. 
Our descriptor is based on a \textit{ResNet18} back-end and a properly parameterized random Gaussian projection layer (RGP). We propose a novel synthetic data generation procedure that enables to make the training procedure extremely efficient and cost-effective when moved to large-scale environments. We also introduce a data normalization/pre-filtering step that helps improve overall performance. Our descriptor is trained to reside in a latent space that maintains as possible a bi-directional mapping with 3D underwater locations. Locally, cosine distances between descriptors aim to encode Euclidean distances between corresponding 3D locations. 
We provide an exhaustive evaluation of the proposed method on both synthetic and real data. Comparisons with state-of-the-art methods show the effectiveness of our method. As a further contribution, we release with this paper an open-source implementation\footnote{\url{https://github.com/ivano-donadi/sdpr}} of our method.


\section{Related Work}
\label{sec:rel_works}

Visual place recognition is a classical computer vision and robotics task \cite{7339473}, which has been developed mainly in the context of optical images acquired with RGB cameras. Initially, traditional vision methods were employed to extract feature descriptors. However, they are limited by the overhead of manual features engineering and are now being gradually supplanted by deep learning-based methods, which on average perform better when properly trained.
In this section, we first present previous works focusing on camera-based place recognition, both leveraging traditional and deep learning-based pipelines and then how they have been adapted in the context of underwater sonar images. 


\subsection{Traditional methods}

Traditional visual place recognition methods can be classified into two categories: global descriptor methods and local descriptor methods. The former approaches primarily focus on the global scene by predefining a set of key points within the image and subsequently converting the local feature descriptors of these key points into a global descriptor during the post-processing stage. For instance, the widely utilized Gist \cite{OLIVA200623} and HoG \cite{1467360} descriptors are frequently employed for place recognition across various contexts. On the other hand, the latter approaches rely on techniques such as Scale Invariant Feature Transformation (SIFT) \cite{790410}, Speeded Up Robust Features (SURF) \cite{10.1007/11744023_32}, and Vector of Local Aggregated Descriptor (VLAD) \cite{5540039} to extract local feature descriptors. Since each image may contain a substantial number of local features, direct image matching suffers from efficiency degradation. To address this, some methods employ bag-of-words (BoW) \cite{1238663} models to partition the feature space (e.g., SIFT and SURF) into a limited number of visual words, thereby enhancing computational efficiency.

\subsection{Deep learning-based methods}

These methods typically rely on features extracted from a backbone CNN, possibly pre-trained on an image classification dataset \cite{10.1145/3065386}. For example, \cite{7353986} directly uses the convolutional feature maps extracted by an AlexNet backbone as image descriptors. Other methods add a trainable aggregation layer to convert these features into a robust and compact representation. Some studies integrate classical techniques such as BoW and VLAD into deep neural networks to further encode deep features as global descriptors, among others in NetBoW \cite{8537942}, NetVLAD \cite{7937898}, and NetFV \cite{miech2018learnable}. Expanding on this, \cite{8099829} introduced the Contextual Re-weighting Network (CRN), which estimates the weight of each local feature from the backbone before feeding it into the NetVLAD layer. Additionally, \cite{8099829} introduced spatial pyramids to incorporate spatial information into NetVLAD.
Moreover, several other studies introduced semantic information to enhance the network's effectiveness by integrating segmentation tasks \cite{rs12233890} and object detection tasks \cite{Snderhauf2015PlaceRW}.

\subsection{Underwater place recognition}

Traditional vision methods, such as SURF \cite{6003474} and BoW technique based on ORB features \cite{10.1007/978-3-031-21065-5_6}, have been used in underwater camera-based place recognition tasks. However, these methods face limitations in complex underwater environments that require laborious pre-processing steps \cite{8869489}. Deep place recognition solutions, such as an attention mechanism that leverages uniform color channels \cite{MaldonadoRamrez2016RoboticVT} and probabilistic frameworks \cite{7404369}, have been proposed to address these challenges. However, the attenuation of electromagnetic waves in water limits their effectiveness.
As a result, underwater sonars have garnered attention from researchers as they are not affected by the aforementioned environmental conditions. For the first time, \cite{7759205} proposed the use of features learned by convolutional neural networks as descriptors for underwater acoustic images. Building upon this, \cite{8260693} introduced a siamese CNN architecture to predict underwater sonar image similarity scores. Then, \cite{8614109} presented a variant of PoseNet that relies on the triplet loss commonly used in face recognition tasks. In particular, an open-source simulator\cite{inproceedings}  was utilized in this work to synthesize forward-looking sonar (FLS) images, from which the network learns features and ultimately performs well on real-world sonar datasets. Inspired by this, \cite{9267885} utilized the triplet loss combined with ResNet to learn the latent spatial metric of side-scan sonar images, enabling accurate underwater place recognition. Additionally, \cite{8793550} employed a fusion voting strategy using convolutional autoencoders to facilitate unsupervised learning of salient underwater landmarks.

\section{Method}
\label{sec:method}

\subsection{Synthetic dataset generation}
\label{sec:synth_data_generation}
\begin{figure}[t]
    \centering
    \begin{subfigure}[b]{0.32\textwidth}
        \centering
        \includegraphics[width=\textwidth]{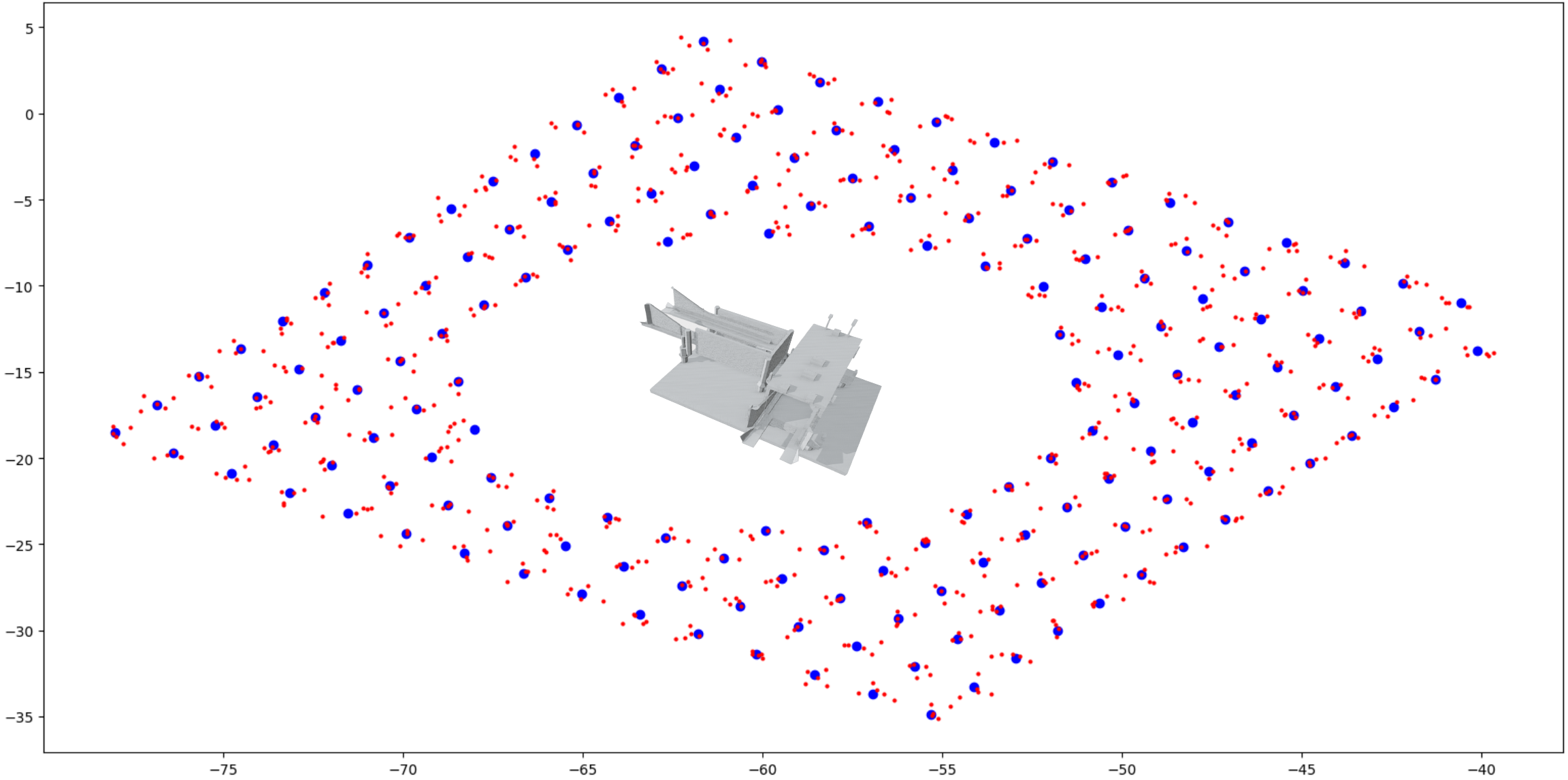}
        \caption{Asset 1.}
        \label{fig:objects:garage}
    \end{subfigure}
    \begin{subfigure}[b]{0.32\textwidth}
        \centering
        \includegraphics[width=\textwidth]{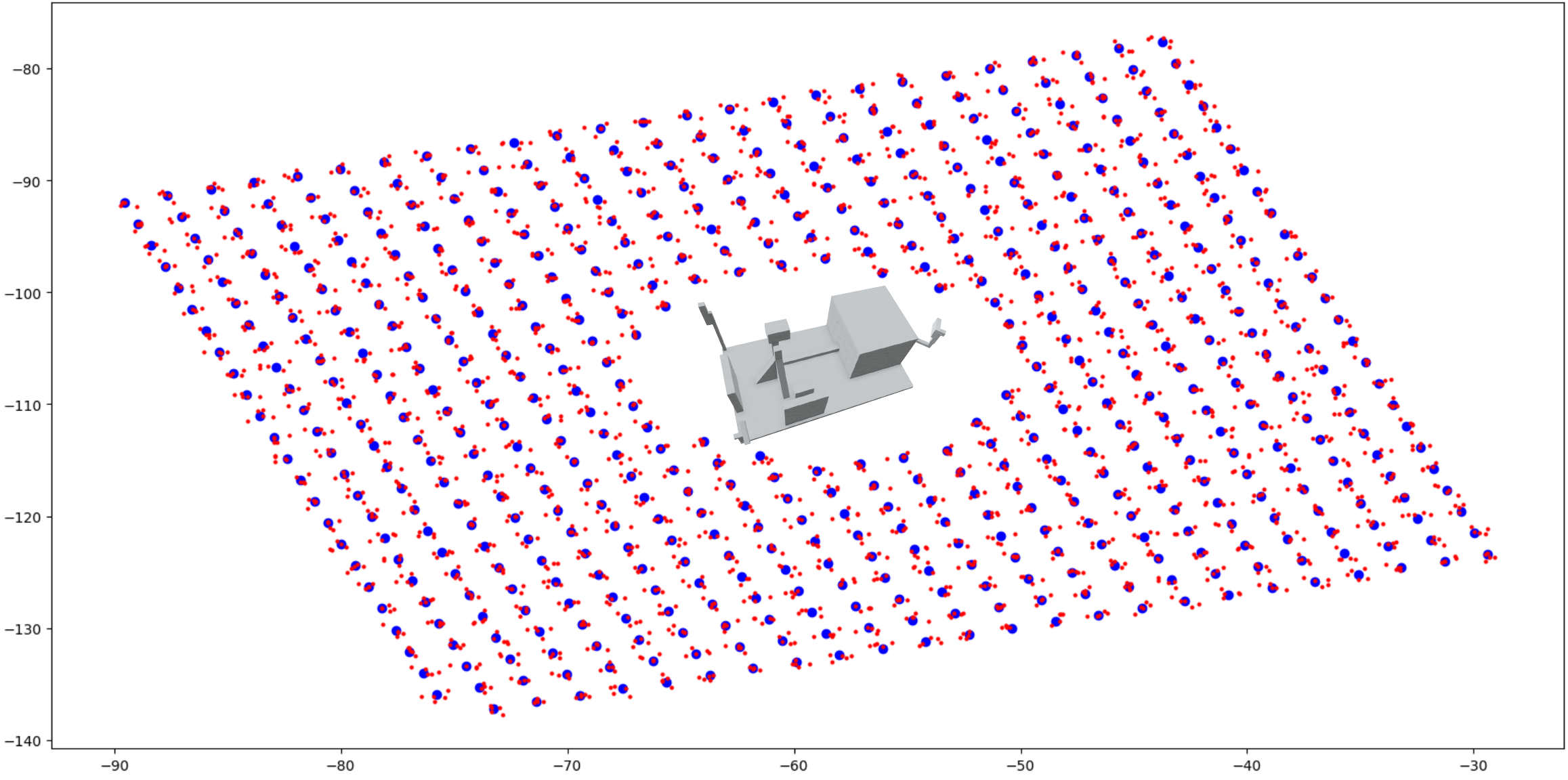}
        \caption{Asset 2.}
        \label{fig:objects:plet}
    \end{subfigure}
    \begin{subfigure}[b]{0.32\textwidth}
        \centering
        \includegraphics[width=\textwidth]{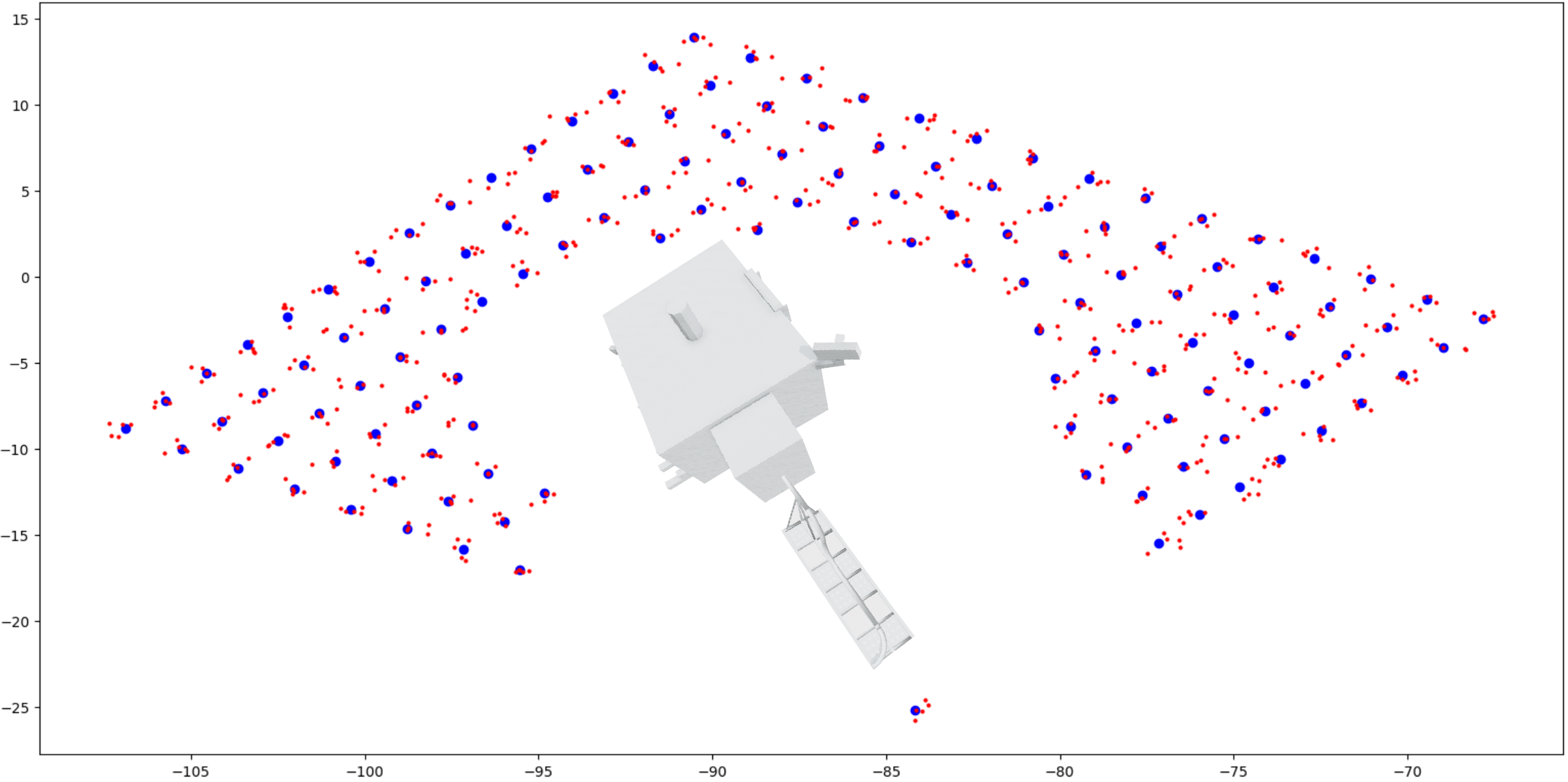}
        \caption{Asset 3.}
        \label{fig:objects:xmastree}
    \end{subfigure}
    \hfill
    \caption{Pose sampling distribution around each of the three objects in the simulated environment. Asset 2 and 3 were used for training the network and asset 1 was used for validation. Blue points are anchor poses and red points are possible positive/negative samples.}
    \label{fig:objects}
    \vspace{-4mm}
\end{figure}

Acquiring and annotating large quantities of sonar images to train and evaluate deep learning models is expensive and time-consuming. Moreover, collecting ground truth pose data can be impractical when the AUV is not close to the water surface since the GPS signal is not available. For this reason, we decided to collect our training data from a simulated environment in which we can freely control sonar parameters and easily retrieve ground truth pose information. In detail, sonar data has been collected using the simulation tool proposed in \cite{zhang2022dave} and the Gazebo simulator \footnote{https://gazebosim.org/home}. A Tritech Gemini 720i sonar sensor with 30m range and 120$^\circ$ horizontal aperture has been simulated for the acquisitions. The data acquisition pipeline was set up according to the following steps:

\begin{itemize}
    \item set up a simulated environment in Gazebo, populated by 3 man-made underwater structures (assets) widely separated in space, and an AUV equipped with the sonar sensor described above 
    (\figref{fig:objects});
    \item define a 2D square grid centered on each object and remove grid cells colliding with the framed object; in our simulations, the grid cell size is 2m$\times$2m and the total grid size is 50m$\times$50m (\figref{fig:objects:garage}, \ref{fig:objects:plet}, \ref{fig:objects:xmastree}). The size of the grid was chosen based on both the range of the sonar and the size of the asset. The cell's size parameters, instead, were chosen based on a trade-off that tried to maximize the difference between anchors' sonar images, while ensuring sufficient fine-grained spacing for localization purposes;
    \item move the AUV at the center of each cell, and change its orientation so that it is facing the object at the center of the grid; then acquire a sonar scan and register it as an anchor (see \secref{sec:triplet-loss});
    \item sample 5 other sonar scans by adding 0-75cm of position noise with respect to the center of the cell, and register them all as possible positives/negatives (see \secref{sec:triplet-loss}).
\end{itemize}

We collect the datasets for the 3 assets separately, obtaining a total of 876 anchors and 4380 possible positives and negatives. With this dataset, we can assess the generalization ability of our method by training it on two assets and validating it on an unseen structure. The underwater assets and pose sampling distribution can be seen in \figref{fig:objects}. 

It should be remarked that to properly assess the generalization capabilities of the proposed approach, the assets used in our simulated environment have a significantly different geometry than the ones framed in the public datasets used for the final experiments (see \secref{sec:datasets}).

\subsection{Triplet descriptor learning}
\label{sec:triplet-loss}

In our approach, we follow a basic framework that is typically exploited for deep descriptor computation (e.g., \cite{Xie2021}). We compute the descriptor of the sonar reading as the embeddings of a convolutional neural network (CNN) fed with the sonar image. The CNN is trained over a large dataset of sonar images by using a triplet loss. Each data item is composed by a reference sonar image (called anchor), a sonar image with a similar field of view (FOV) to the anchor image (called positive), and an image with a significantly different FOV than the anchor (called negative). Such a strategy has already been applied to sonar images in works such as \cite{8614109}, \cite{8260693}. Let $A$, $P$, and $N$ be descriptors for respectively the anchor, positive and negative images; the triplet loss is then defined as:

\begin{equation}
    L_{T} = max\{0, d(A,P) - d(A,N) + m\}
\end{equation}
where $d$ is a distance metric (usually the Euclidean distance) between descriptors and $m$ is the margin hyper-parameter, representing the desired difference between positive-anchor similarity and negative-anchor similarity. A  graphical intuition of the triplet loss is provided in \figref{fig:sonar:triplet}. In our implementation the distance metric used is based on cosine similarity rather than Euclidean distance: let $C_S(x,y) = \frac{x \cdot y}{||x||_2\cdot||y||_2}$ be the cosine similarity between two vectors $x$ and $y$, then the distance metric between $x$ and $y$ is defined as $d_S(x,y) = 1 - C_S(x,y)$. When vectors are constrained to unit length, the squared Euclidean distance and the cosine distance are proportional, and we chose to use the latter because it provides a more intuitive choice of margin. 

\subsection{Triplet generation}
\label{sec:triplet_generation}

\begin{figure} [t]
    \centering
    \begin{subfigure}[b]{0.4\textwidth}
        \centering
        \includegraphics[width=\textwidth]{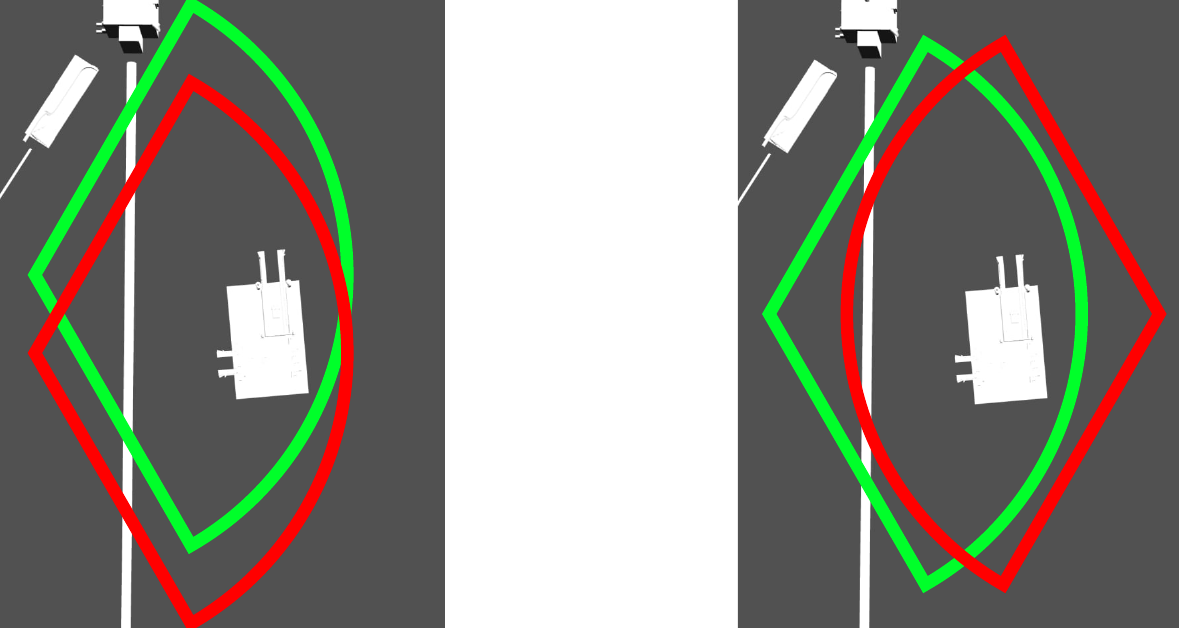}
        \caption{}
        \label{fig:sonar:fov}
    \end{subfigure}
    \hfill
    \begin{subfigure}[b]{0.4\textwidth}
        \centering
        \includegraphics[width=\textwidth]{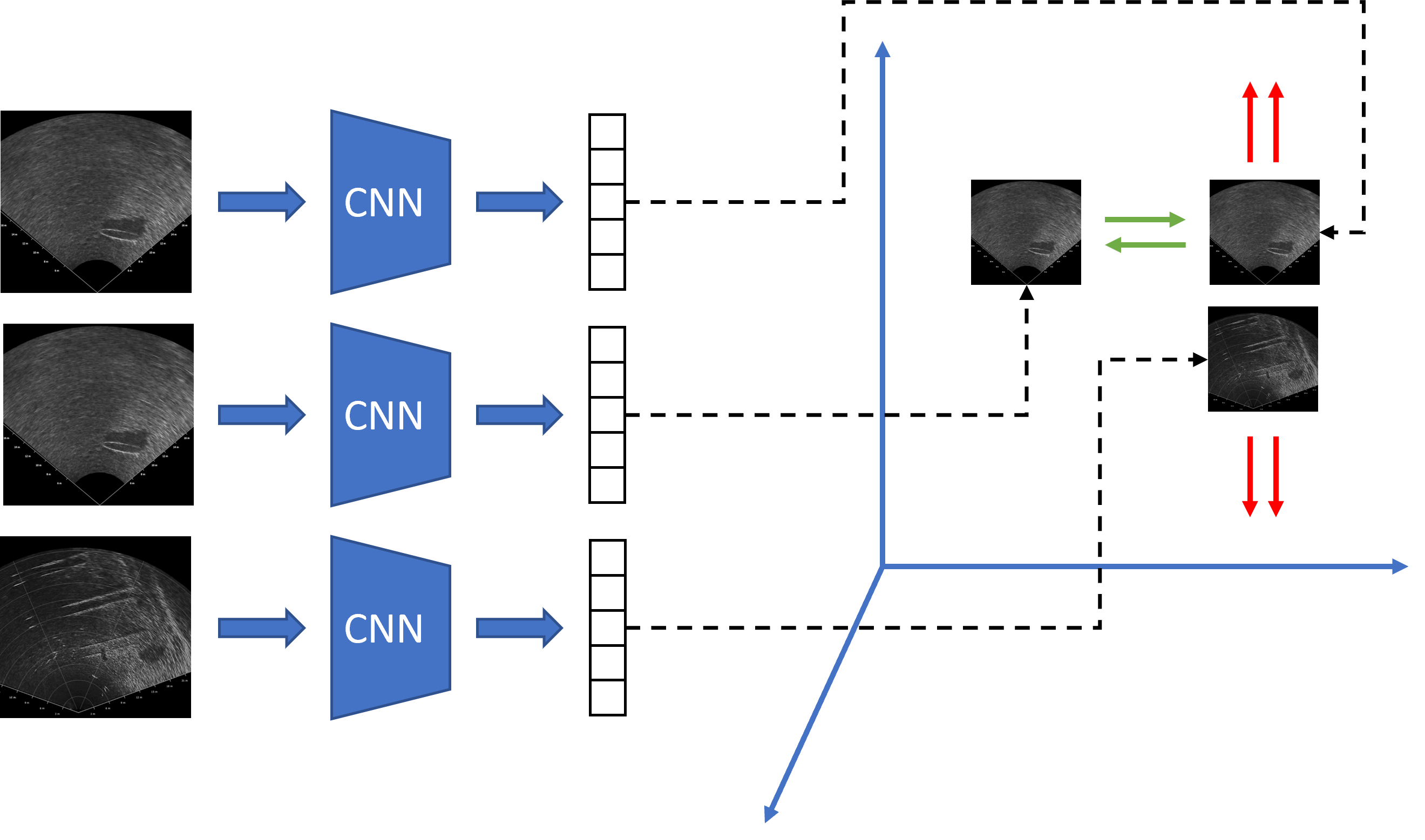}
        \caption{}
        \label{fig:sonar:triplet}
    \end{subfigure}
    \caption{\subfigref{fig:sonar:fov} shows two different cases of large sonar FOV overlap: the pair on the left should be classified as framing the same location, while the one on the right shouldn't, since the two objects sides are not symmetrical. \subfigref{fig:sonar:triplet} provides a graphical intuition of the triplet loss training. This loss is used to learn a descriptor that makes similar images spatially close (e.g., top and middle images) and different images spatially distant (e.g., top and bottom images).}
    \label{fig:sonar}
\end{figure}

The triplet loss requires classifying training samples as positives or negatives with respect the any given anchor. To do so, we employed the metric proposed in \cite{8260693}, in which the similarity between two sonar scans is defined as the relative area overlap between the two sonar FOVs. Given a threshold $\tau$, we classify a sample as negative if its similarity w.r.t the anchor is lower than the threshold, and as positive otherwise. Due to the grid-like nature of our training dataset, additional care is needed when computing sonar similarity scores: in particular, we need to set a maximum allowed orientation difference between the two scans in order to give low similarity to pairs at the opposite ends of the framed object, which share a large amount of area but frame two different object geometries (see \figref{fig:sonar:fov}). In order to provide consistently significant samples for each anchor at each epoch, we sample a set of $n_{neg}$ possible negative samples and $n_{pos}$ possible positive samples, and perform batch-hardest negative mining \cite{hermans2017defense} and batch-easiest positive mining \cite{xuan2020improved} inside this sub-set by selecting the negative and positive samples whose descriptors are the closest to the anchor's, speeding up the training process while allowing a certain degree of intra-class variance. The actual values of the parameters we used in our experiment are detailed in \tabref{tab:params}. In particular, $n_{pos}$ was chosen as to include all points sampled close to an anchor, as described in \secref{sec:synth_data_generation}, while the choice of $n_{neg}$ was upper-bounded by memory constraints.

\subsection{Network structure}
\label{sec:network}

\begin{figure}[t]
    \centering
    \includegraphics[width=\textwidth]{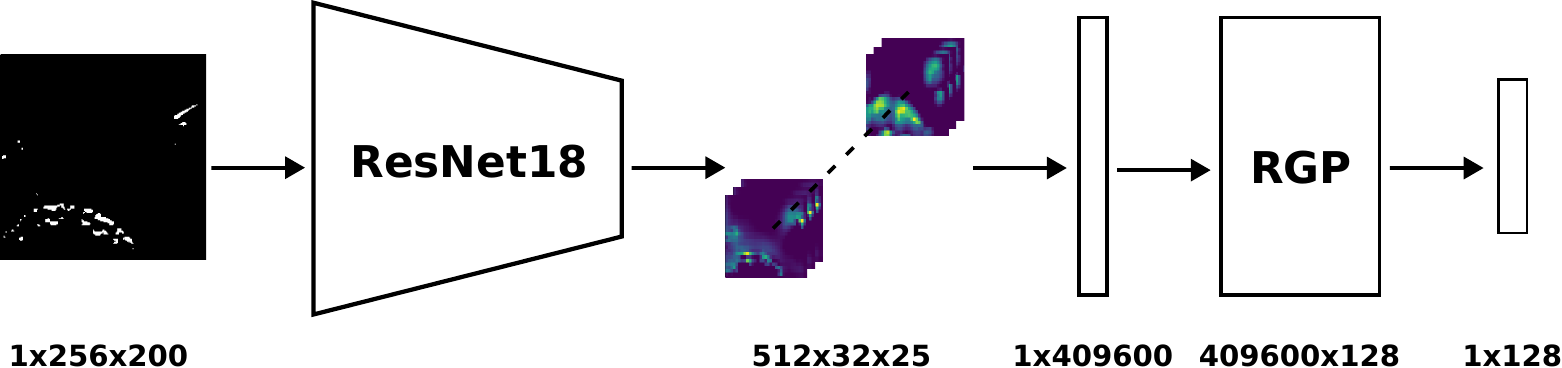}
    \caption{Graphical representation of our descriptor extraction network.}
    \label{fig:network}
\vspace{-5mm}
\end{figure}

The backbone of our model is a lightweight ResNet18 encoder network, modified to take single-channel inputs, with an output stride of 32 and without the final average pooling and fully connected layers. The network's input is raw sonar images (beams$\times$bins) resized to 256$\times$200. Starting from the input image, we extract, in the final convolutional layer, 512 feature maps of size 25$\times$32, which are then flattened in a single high-dimensional vector descriptor. To reduce the descriptor size, we employ random Gaussian projection (RGP), as in \cite{Zaffar_2021}, obtaining 128-dimensional image descriptors. Finally, descriptors are normalized to unit length. Our experiments reported no significant advantage when increasing the descriptor size for this task. Overall, our architecture and training strategy are similar to \cite{9267885}, with three key differences: we use cosine distance in place of Euclidean distance in the triplet loss, we replace the last fully connected layer with a fixed RGP matrix (resulting in fewer trainable parameters), and we focus on forward-looking sonar and not on side-scan sonar. A graphical representation of this pipeline can be found in \figref{fig:network}.
We also developed a different network structure based on NetVLAD descriptors \cite{7937898}, which are extensively used for camera-based place recognition, by keeping the same backbone and substituting the RGP layer with a NetVLAD aggregation layer. The performance of both methods is compared in \secref{sec:experiments}.

\begin{table}[t]
    \centering
    \begin{tabular}{| c | c | c | c | c | }
        \cline{1-2}
        \cline{4-5}
        Parameter & Value && Parameter & Value\\
        \cline{1-2}
        \cline{4-5}
        $\tau$ & $0.7$ && $m$ & $0.5$\\
        \cline{1-2}
        \cline{4-5}
        $n_{neg}$ & $10$ && $n_{pos}$ & $5$\\
        \cline{1-2}
        \cline{4-5}
        $n_w$ & $40$ && $P_{fa}$ & $0.1$\\
        \cline{1-2}
        \cline{4-5}
    \end{tabular}
    \caption{List of parameter values used in our experiments.}
    \label{tab:params}
    \vspace{-5mm}
\end{table}
\subsection{Image enhancement}
\begin{figure}[ht]
 \vspace{-5mm}
     \centering
     \includegraphics[width=\linewidth]{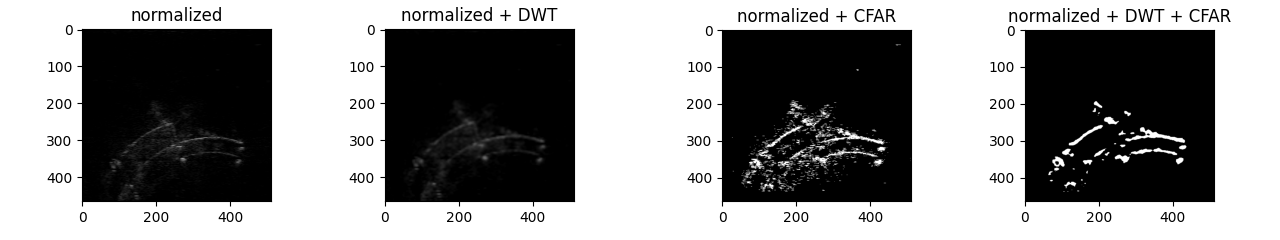}
     \caption{Example of the proposed image enhancing procedure on a scan from the Wang2022 dataset \cite{wang2022virtual}. Notice how applying CFAR directly on the normalized image is not enough to differentiate between objects of interest and noise.}
     \label{fig:filtering}
 \vspace{-10mm}
 \end{figure}

 Sonar images taken in a real environment are known to be plagued by a significant amount of additive and multiplicative noise \cite{8492877}. Furthermore, the sonar emitter might not provide a uniform insonification of the environment, resulting in an intensity bias in some image regions. 
 To mitigate these problems, we enhance real sonar images in three steps: first, we normalize the images to uniform insonification \cite{dosSantos2018} \cite{kim2005}, then we filter them using discrete wavelet transforms (DWTs) as in \cite{7557212}, and finally we threshold the images with a Constant False Alarm Rate (CFAR) thresholding technique, obtaining binary images \cite{wang2022virtual} \cite{7185}. In particular, in the first step, the sonar insonification pattern is obtained by averaging a large number of images in the same dataset, and the resulting pattern is used to normalize pixel intensities. In the last step, synthetic images were thresholded using GOCA-CFAR (greatest of cell averaging) and real images using SOCA-CFAR (smallest of cell averaging). Both methods scan each sonar beam independently and select a personalized threshold for each sonar cell in the beam. Let $c$ be the current sonar cell, and let $w_l$ and $w_t$ be windows of $n_w$ cells respectively leading and trailing $c$ in the beam. Let $\overline{w_l}$ and $\overline{w_t}$ be the average intensity in the two windows, then the threshold for SOCA-CFAR is selected as min$(\overline{w_l}, \overline{w_t})$, while the threshold for GOCA-CFAR is equal to max$(\overline{w_l}, \overline{w_t})$. Both thresholds are then weighted by a factor dependent on the desired false alarm rate $P_{fa}$. All CFAR parameters were selected by maximizing the human-perceived quality of the sonar images. Our DWT filtering procedure is identical to \cite{7557212}. An example of the enhanced image can be seen in \figref{fig:filtering}.

\section{Experiments}
\label{sec:experiments}

\subsection{Datasets}
\label{sec:datasets}

\begin{figure}
    \centering
    \begin{subfigure}[t]{0.46\textwidth}
        \centering
        \includegraphics[width=\textwidth]{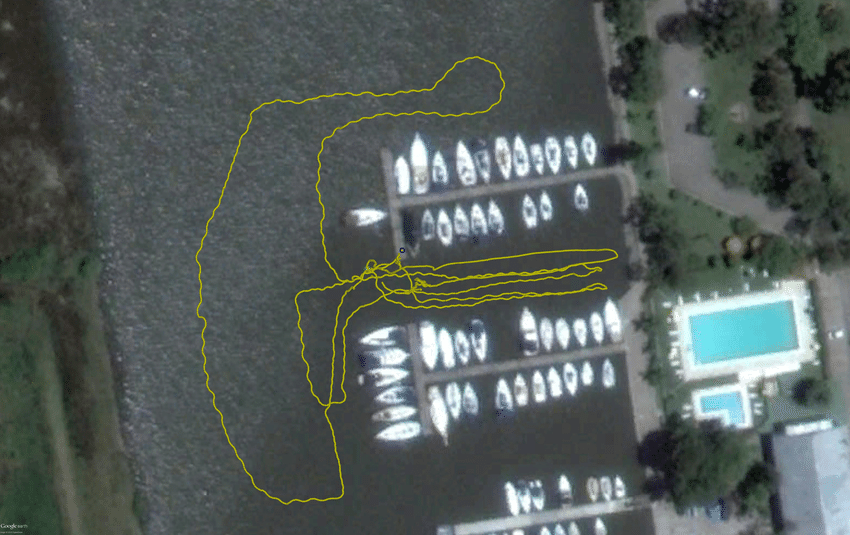}
       \caption{}
        \label{fig:trajectory:2014}
    \end{subfigure}
    \hfill
    \begin{subfigure}[t]{0.35\textwidth}
        \centering
        \includegraphics[width=\textwidth]{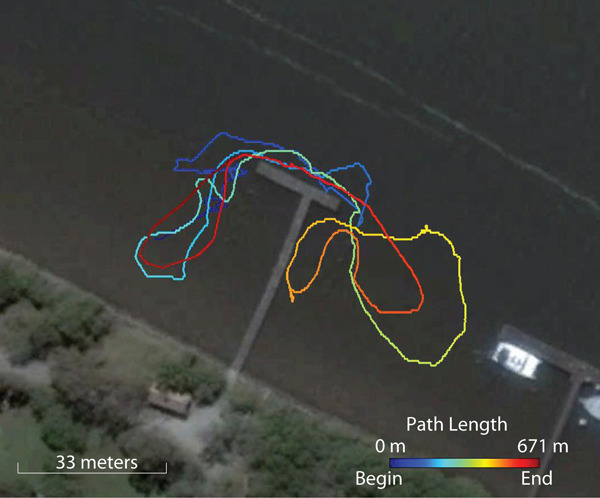}
        \caption{}
        \label{fig:trajectory:2017}
    \end{subfigure}
    \vspace{-4mm}
    \caption{AUV trajectories for the two Aracati datasets: (a) Aracati2014 \cite{Machado2016}. (b) Aracati2017 \cite{dosSantos2018}.}
    \label{fig:trajectory}
\end{figure}

We validated our method in three publicly available datasets:

\begin{itemize}
    \item \textbf{Wang2022} \cite{wang2022virtual}: This dataset was collected on a real dock environment to test the sonar-based SLAM approach presented in \cite{wang2022virtual}, using a BlueROV AUV equipped with an Oculus M750d imaging sonar, using maximum range equal to 30m and a horizontal aperture of 130$^\circ$. This dataset has no ground truth pose annotations, so we relied on the author's SLAM implementation to compute an accurate trajectory for the AUV. Unfortunately, pose annotations extracted with this method are only available for keyframes, resulting in 187 total annotated sonar scans, of which only 87 contain actual structures.
    
    \item \textbf{Aracati2014} \cite{SILVEIRA2015212}: This dataset was acquired in the Yacht Club of Rio Grande, in Brazil with an underwater robot mounting a Blueview P900-130 imaging sonar with a maximum range equal to 30m and a horizontal aperture of 130$^\circ$. Ground truth pose data was acquired by attaching a GPS system to a surfboard connected to the robot. This dataset contains over 10k sonar images, of which 3675 can be synchronized with the position and heading sensors. An additional filter is required to discard sonar scans looking at an empty scene, such as when the AUV is traveling away from any object, so 1895 images are actually usable for testing, similarly to \cite{dosSantos2018}. The AUV's trajectory is displayed in \figref{fig:trajectory:2014}. 

    \item \textbf{Aracati2017} \cite{dosSantos2018}: This dataset was acquired in the same location as Aracati2014, but performed a different trajectory. The sonar sensor and parameters are unchanged except for the maximum range, which was set to 50m. 14350 annotated sonar images are provided, of which 8364 contain some underwater structures. The AUV trajectory can be seen in \figref{fig:trajectory:2017}.
\end{itemize}
\vspace{-6mm}
\vspace{-10mm}
\subsection{Metrics}

Given a trained descriptor extractor $D$, it is possible to match two sonar images $x,y$ by computing the distance between their descriptors $d_S(D(x),D(y))$ and comparing it against a certain threshold. Image pairs are considered a positive match if their descriptor distance falls below the threshold, and a negative match otherwise. When considering all possible image pairs in a dataset, it is possible to construct a table of all positive and negative matches, which is dependent on the value of the threshold used to classify pairs as positive or negative matches. An analogous table can be computed by considering ground truth matches based on the metric described in \secref{sec:triplet_generation}. Comparing the two tables it is possible to extract true positives (TP), true negatives (TN), false positives (FP), and false negatives (FN) as in any binary classification problem. Based on these values, we evaluated our model using several metrics commonly used in visual/sonar place recognition:

\begin{itemize}
    \item \textbf{Area under the precision/recall curve (AUC)}: given a set of values for the matching threshold, we can compute precision ($\frac{TP}{TP+FP}$) and recall ($\frac{TP}{TP+FN}$) values for each threshold. The precision/recall curve is then obtained by using the precision as the y-axis and the recall as the x-axis. A high area under the curve metric signifies both high precision and high recall. We also report the precision and recall values at the optimal threshold, which is the one maximizing the F1 score.
    
    \item \textbf{Recall at 95\% precision (R@95P)}: if a visual place recognition algorithm is to be used inside a loop detection system for SLAM, it is necessary to have high precision to avoid wrong loop closures. This metric allows to see the percentage of loop closures detected when requiring high precision. 
    
    \item \textbf{Precision over FOV overlap}: In this metric, used by \cite{8614109}, we compute the FOV overlap between the test sonar image and its nearest neighbor - in the descriptors space - in the dataset, and compare it against a set of FOV overlap thresholds (from $10\%$ to $90\%$ in steps of $10\%$), obtaining the percentage of dataset images whose nearest neighbor share a percentage of FOV over the threshold. As in \cite{8614109}, we also report this metric when removing a window of $s=3$ seconds, leading and trailing the query image in the trajectory, from the nearest neighbor search: this allows evaluating the ability of the sonar matching method to recognize previously seen locations when circling back to them, rather than in the subsequent trajectory frames. 
\end{itemize}

\subsection{Results}

In this section, we compare the effectiveness of our model against the NetVLAD-based one on all three real datasets of \secref{sec:datasets}. For the Aracati2014 dataset, we also provide a comparison with another triplet-loss-based sonar place recognition network, introduced in \cite{8614109}, which we will refer to as \methodRibeiro{} in the remainder of the paper. Results, in this case, are taken directly from \cite{8614109}. This particular method was trained on the real Aracati2017 dataset and tested on Aracati2014, using 2048-dimensional descriptors. Additionally, we report the performance of our model with randomly initialized weights to assess the contribution of our training strategy to the network's performance. We will refer to this network as \methodrandom{} in the remainder of the paper.

\subsubsection{Validation:}
\begin{figure}[h]
    \centering
    \begin{subfigure}[b]{0.35\textwidth}
        \centering
        \includegraphics[width=\textwidth, height=4cm, trim={0.3cm 0.5cm 1cm 1cm}, clip]{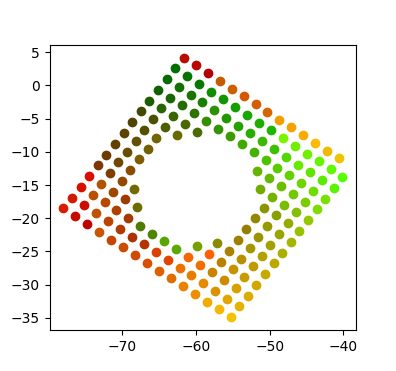}
        \caption{}
        \label{fig:tsne:ours}
    \end{subfigure}
    \hfill
    \begin{subfigure}[b]{0.35\textwidth}
        \centering
        \includegraphics[width=\textwidth, height=4cm]{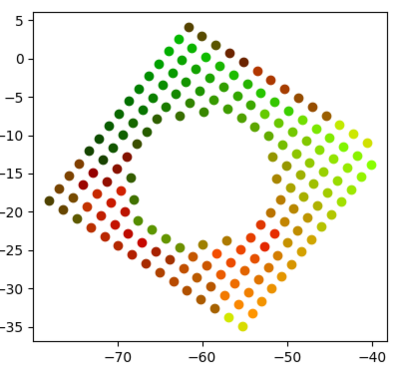}
        \caption{}
        \label{fig:tsne:vlad}
    \end{subfigure}
    \vspace{-3mm}
    \caption{Descriptor distribution in the Euclidean space: the color of each anchor is obtained by exploiting the 2D t-SNE projection of its descriptor as red and green RBG components, thus, the more smooth is the change in color, the better the descriptor distribution. In \subfigref{fig:tsne:ours} the results are obtained using our approach, while in  \subfigref{fig:tsne:vlad} are obtained using the \methodvlad{} method.}
    \label{fig:tsne}
\end{figure}

Our training pipeline, outlined in \secref{sec:method}, aims to establish a mapping between the environment's 3D space and the 128-dimensional latent descriptor space. To evaluate the effectiveness of our approach, we used a validation dataset consisting of regularly spaced anchors in the 3D space (\figref{fig:objects:garage}). We expect this regularity to be reflected in the latent space such that the difference between neighboring anchor descriptors should remain constant across the whole dataset. The qualitative results reported in \figref{fig:tsne} show that our approach successfully achieved this goal, with smoother transitions between nearby locations than NetVLAD.
\begin{figure}[h]
\vspace{-5mm}
    \centering
    \includegraphics[width=\textwidth]{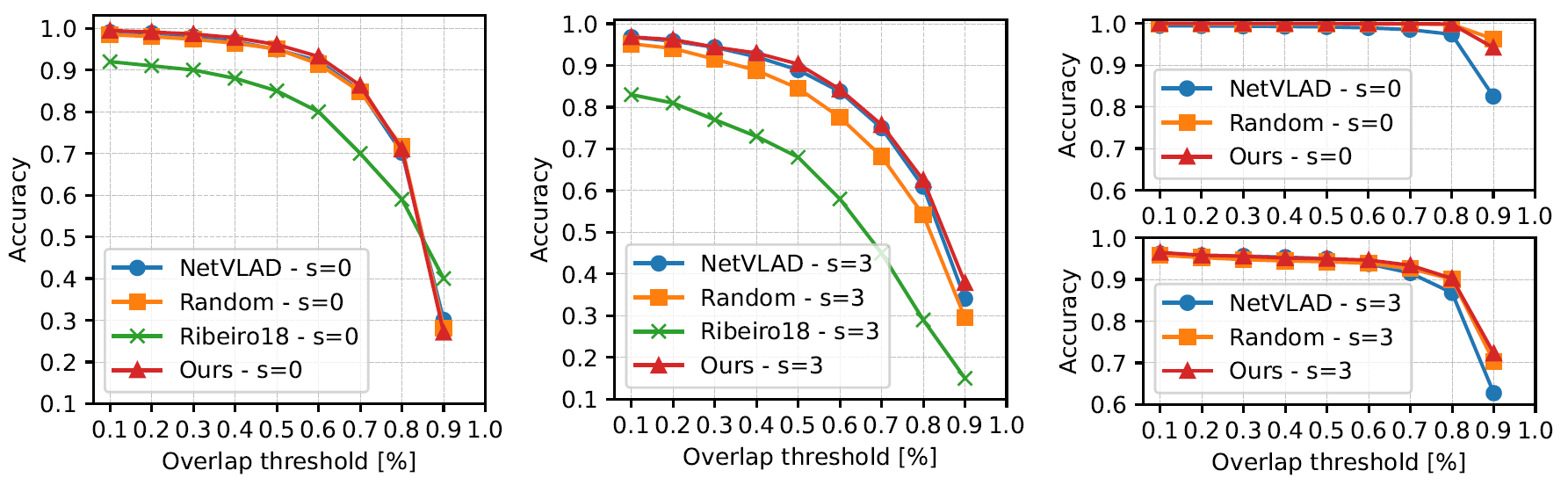}
    \caption{Evaluation results on both Aracati datasets. In the first two columns we report the results for all methods on the Aracati2014 dataset, while the last column contains the accuracy over FOV overlap on the Aracati2017 dataset.}
    \label{fig:prec_over_overlap:2014-2017}
    \vspace{-5mm}
\end{figure}
\begin{table}[h]
    \centering
    \begin{tabular}{|c|c||c|c|c|c||c|c|c|c||c|c|c|c|}
        \hline
         \multicolumn{2}{|c||}{\multirow{2}{*}{Method}} & \multicolumn{4}{c||}{Aracati2014} & \multicolumn{4}{c||}{Aracati2017} & \multicolumn{4}{c|}{Wang2022} \\
        \cline{3-14}
         \multicolumn{2}{|c||}{} & AUC & P & R & R@95P & AUC & P & R & R@95P & AUC & P & R & R@95P \\
        \hline
        \hline
        \multirow{3}{*}{\STAB{\rotatebox[origin=c]{90}{$s=0$}}}
        & \methodours{} & \textbf{.86} & \textbf{.95 }& \textbf{.95} & \textbf{.93} & .75 & \textbf{.99} & \textbf{.99} & \textbf{.99} & .65 & .89 & .80 & \textbf{.67}\\
        \cline{2-14}
        & \methodvlad{} & .83 & .92 & .92 & .14 & .68 & .95  & .94 & .00 & \textbf{.79}& \textbf{.90} & \textbf{.86} & .65\\
        \cline{2-14}
        & \methodrandom{} & .85 & .94 & .93 & .78 & \textbf{.91} & .99 & .99 & .99 & .72 & .90 & .73 & .53\\
        \hline
        \hline
        \multirow{3}{*}{\STAB{\rotatebox[origin=c]{90}{$s=3$}}}
        & \methodours{} & \textbf{.75} & \textbf{.86} & \textbf{.84} & .00 & .86 & \textbf{.93} & \textbf{.93} & \textbf{.83} & - & - & - & - \\
        \cline{2-14}
        & \methodvlad{} & .72 & .83 & .83 & \textbf{.01} & .70 & .90  & .89 & .00 & - & - & - & -\\
        \cline{2-14}
        & \methodrandom{} & .63 & .79 & .77 & .00  & \textbf{.88} & .92 & .92 & .80 & - & - & - & - \\
        \hline
    \end{tabular}
    \caption{Network performance. P and R columns contain precision and recall values at the optimal matching threshold. Due to the limited size of the Wang2022 dataset, we do not report the results for $s=3$.}
    \label{tab:results}
\vspace{-8mm}
\end{table}
\subsubsection{Real datasets:} \tabref{tab:results} reports the quantitative evaluation results of both our method and the NetVLAD-based network regarding the area under the precision/recall curve and the recall at 95\% precision. We can see that our method is the best-performing one on both Aracati datasets, while NetVLAD performs best on the Wang2022 dataset. It might seem surprising that the non-trained network has a performance so close to the other two methods, especially in the case of $s=0$, but it is justified by the fact that semi-identical images will provide semi-identical responses to any stable filtering technique, even if random. However, given the huge shift in the domain between training and test datasets, the fact that the trained models still provide superior place recognition abilities proves the effectiveness of the training. In \figref{fig:prec_over_overlap:2014-2017},  we report the precision over FOV overlap for both our models and the \methodrandom{} and \methodRibeiro{} models. The data for the latter method was taken directly from the original paper. It is possible to see how, in the case $s=0$, even the non-trained network is able to correctly retrieve matching images simply by matching subsequent (almost identical) frames. When setting $s=3$, instead, the two trained networks show their learned ability to match the same scene from different points of view. 

\vspace{-3mm}
\section{Conclusions}
\label{sec:conclusions}
In this paper, we proposed a compact sonar image descriptor for underwater place recognition that is computed using deep CNNs trained only on simulated data of underwater scenarios. We demonstrate the effective generalization capabilities of our descriptor through extensive experiments and evaluations. In particular, our sonar descriptor has been validated both on simulated and real data and tested against other recent state-of-the-art approaches, obtaining promising results.\\
Future developments include the integration of a module capable of determining the distinctiveness of locations, in order to automatically remove sonar images that belong to empty areas, a domain-adaptation strategy to better mimic real data in simulation, and the use of more advanced sonar simulators \cite{Potokar22icra}.

\vspace{-2mm}
\subsubsection{Acknowledgements}
This work was supported by the University of Padova under Grant UNI-IMPRESA-2020-SubEye.
%
%
%
\bibliographystyle{splncs04}
\bibliography{references}

\end{document}